\documentclass[10pt]{article}
\usepackage{spconf,amsmath,graphicx}
\usepackage[pass,top=0.75in,bottom=1.1in,left=0.75in,right=0.75in]{geometry}
\usepackage{enumitem}
\usepackage{subcaption}
\setlist{nosep,leftmargin=14pt}

\title{A Multimodal Slice Discovery Framework for Systematic Failure Detection and Explanation in Medical Image Classification}

\name{Yixuan Liu$^{1}$, Kanwal K. Bhatia$^{2}$, Ahmed E. Fetit$^{1}$}
\address{$^{1}$Department of Computing, Imperial College London, UK\hspace{3mm}$^{2}$Aival, London, UK
}

\begin{document}
\maketitle
\begin{abstract}
Despite advances in machine learning-based medical image classifiers, the safety and reliability of these systems remain major concerns in practical settings. Existing auditing approaches mainly rely on unimodal features or metadata-based subgroup analyses, which are limited in interpretability and often fail to capture hidden systematic failures.
To address these limitations, we introduce the first automated auditing framework that extends slice discovery methods to multimodal representations specifically for medical applications. Comprehensive experiments were conducted under common failure scenarios using the MIMIC-CXR-JPG dataset, demonstrating the framework’s strong capability in both failure discovery and explanation generation. Our results also show that multimodal information generally allows more comprehensive and effective auditing of classifiers, while unimodal variants beyond image-only inputs exhibit strong potential in scenarios where resources are constrained.

\end{abstract}
\section{Introduction}
\label{sec:intro}

In recent years, the performance of medical image classification models has improved substantially, but their reliability and safety remain major concerns in practical use due to challenges such as fairness issues, spurious correlations, and limited domain generalization.
These systems can still fail on specific subgroups.
Although many ongoing studies attempt to minimize such risks during model design, continuous monitoring and auditing remain essential responsibilities for both developers and end users.

Traditional model auditing typically relies on subgroup analysis based on metadata~\cite{bissoto2025subgroup}, which is often unavailable and fails to capture errors beyond predefined subgroups.
Recently, \textit{Slice Discovery Methods (SDMs)}~\cite{eyuboglu2022domino} have been proposed to automatically identify failure-prone subgroups with less dependence on metadata, offering a more flexible and data-driven approach to model auditing.
These techniques were originally developed for vision-only tasks. Although several studies~\cite{bissoto2025subgroup, olesen2024slicing} have adapted SDMs to medical imaging, they remain limited to image inputs and overlook the multimodal nature of clinical data. Moreover, their failure interpretations are largely descriptive and rely on manual inspection, as modalities beyond images remain underexplored.

.
To address these limitations, this work introduces a fully automated multimodal auditing framework for medical image classifiers in black-box settings.
The pipeline acts as an independent third-party auditor that identifies where a classifier systematically  fails, without requiring access to model internals, training data, or costly expert annotations.
By incorporating the multimodal nature of healthcare data, the framework extends slice discovery methods beyond vision-only settings.
The multimodal representation improves the detection of systematic failures and supports automatic, clinically meaningful explanations: bridging model auditing and real-world medical interpretation. To the best of our knowledge, this is the first novel extension of SDMs to multimodal embeddings, combining image, report, and metadata representations for more reliable and interpretable medical auditing.

\section{Method}

\subsection{Problem Formulation}

Let \( h_\theta \) denote the black-box image classifier to be audited on a multimodal dataset 
\( D = \{(x_i, y_i, z_i)\}_{i=1}^N \), 
where \( x_i \in \mathcal{X} \) is the image input, \( y_i \in \mathcal{Y} \) is the ground truth label, and \( z_i \in \mathcal{Z} \) represents complementary information such as reports or metadata. 
The classifier produces predictions \(\hat{y}_i = h_\theta(x_i)\).

The framework focuses on detecting and explaining \textit{systematic failures}, where the model consistently underperforms on a coherent subset of the data, referred to as an \textbf{error slice}. 
Samples within a slice share visually observable or semantically meaningful \textbf{attributes} (e.g., acquisition view or patient demographics), and a single sample may belong to multiple slices. 
An \textbf{error attribute} is defined when the model performs significantly worse on samples with that attribute:
\begin{equation}
    e(S_{y, \text{attr}}) \gg e(S_{y, \neg\text{attr}}),
\end{equation}
where \(e(\cdot)\) denotes the error rate on a particular subset \(S\). 
The goal of the auditing pipeline is to automatically identify such error attributes that correspond to systematic failures of the classifier under audit. 
To this end, the framework first discovers potential error slices and then summarizes the underlying error-related features based on the identified slices.

\subsection{Error Identification}
In this stage, we build upon the DOMINO algorithm~\cite{eyuboglu2022domino}, originally developed for image-only inputs, and extend it to a multimodal setting as part of the proposed auditing framework. A Gaussian Mixture Model (GMM) is applied to the joint space of multimodal embeddings $u_i$, ground-truth labels $y_i$, and model predictions $\hat{y}_i$ to identify clusters of semantically coherent samples with high error rates. To simplify the task in black-box setting, we reformulate the multiclass problem into a binary classification task, where target-class samples are labeled as 1 and all others as 0.
The GMM parameters $\phi$ are optimized by maximizing the following likelihood:
\begin{align}
\mathcal{L}(\phi) 
&= \sum_{i=1}^{N}
\log
\sum_{j=1}^{K}
P(S^{(j)} = 1)
P(u_i \mid S^{(j)} = 1) \nonumber \\
&\quad \times 
\big[
P(y_i \mid S^{(j)} = 1)
P(\hat{y}_i \mid S^{(j)} = 1)
\big]^{\gamma},
\label{eq:auditing_loss}
\end{align}
where $\gamma$ balances the cluster error rate and cluster coherence, and $S^{(j)}$ denotes the $j$-th slice. A probability threshold $\beta$ is applied, and only samples with $P(x_i \mid S_j) > \beta$ are assigned to slice $S_j$ for later interpretation and analysis.

To construct the multimodal $u_i$, we build a unified embedding that integrates complementary information from all available modalities. The same multimodal model is used to obtain embeddings for all modalities to ensure representational consistency. Image and text features are directly extracted from the multimodal model, while tabular DICOM metadata is converted into short textual descriptions and encoded through its text encoder. As the classifiers under audit are treated as black boxes, we do not assume any prior knowledge or modality-specific importance. Therefore, embeddings from all modalities are concatenated with equal weights. Concatenation is adopted for its simplicity and completeness, followed by Principal Component Analysis (PCA) for dimensionality reduction before clustering, as PCA effectively preserves structural similarity among samples while improving computational efficiency.

\subsection{Explanation Generation}
\label{token analysis}

To provide interpretable explanations for the discovered error slices, 
we develop a token-based analysis module that identifies textual attributes uderlying.
This module is based on Term Frequency–Inverse Document Frequency (TF-IDF)~\cite{jones1972statistical}, 
chosen for its simplicity and interpretability.
Tokens refer to words extracted from clinical reports or metadata text, 
and the key idea is to find tokens that appear more frequently in misclassified samples. 

To avoid capturing class-specific patterns, 
we construct a \textbf{reference slice} $S_{\text{ref}}$ 
from correctly predicted samples of the same class as the \textbf{error slice} $S_{\text{err}}$. 
For each token $t$, we define the \textbf{distinctiveness score (DS)} as:
\begin{equation}
\text{DS}(t) = \mu_{\text{err}}(t) - \mu_{\text{ref}}(t),
\end{equation}
where $\mu_{\text{err}}(t)$ and $\mu_{\text{ref}}(t)$ denote the mean TF-IDF of $t$ 
in $S_{\text{err}}$ and $S_{\text{ref}}$ respectively. 
Tokens with higher $\text{DS}(t)$ are interpreted as potential error attributes or correlated subgroups. 

To assess the validity of the identified attributes, 
we integrate a multimodal similarity metric into our pipeline. Following the idea of the CLIP Score~\cite{kim2024discovering}, the similarity is computed between each attribute and the corresponding image slices:

\begin{equation}
\begin{split}
\mathrm{sim}(attr, S) &=
\frac{1}{|S|}
\sum_{x \in S}
\bigl\langle f_{\mathrm{img}}(x),\, f_{\mathrm{txt}}(attr) \bigr\rangle,\\
r_{\mathrm{attr}} &=
\mathrm{sim}\bigl(attr, S_{\mathrm{err}}\bigr)
-
\mathrm{sim}\bigl(attr, S_{\mathrm{ref}}\bigr).
\end{split}
\end{equation}

Here, $S_{\mathrm{err}}$ and $S_{\mathrm{ref}}$ 
are the error and reference slices, and $f_{\mathrm{img}}$, $f_{\mathrm{txt}}$ 
are the image and text encoders from the same multimodal model used earlier. A higher $r_{\mathrm{attr}}$ indicates that the token is more closely associated with the systematic-error pattern within the error slice, suggesting that it reflects a potential cause or contributing factor of the model’s failure.

\section{Experiments}

\subsection{Controlled Bias Simulation}
\label{failure}
To systematically evaluate the proposed pipeline, we simulated predefined failure types by introducing attribute-based biases during model training, and tested whether the pipeline could detect and explain them. 
Following Oakden-Rayner et al.\ and Eyuboglu et al.~\cite{oakden2020hidden, eyuboglu2022domino}, three common bias types were considered: 
(1) \textit{Spurious Correlation}, 
(2) \textit{Rare Slice Undertraining}, and 
(3) \textit{Noisy Label Injection}. 
The underperforming group was defined by target label $Y$ and attribute $attr$, and bias was introduced by controlling the quality and proportion of training data $D_\text{train}$. 
The strength of correlation, noise, and rarity were quantified as:
\begin{equation}
\begin{aligned}
\rho(Y, attr) &= \frac{\mathrm{Cov}(Y, attr)}{\sigma_Y \sigma_{attr}}, \\
\text{NoiseRate}_\alpha &= \frac{\# \text{ flipped labels in } D_{Y, attr}}{|D_{Y, attr}|}, \\
R(Y, attr) &= \frac{|D_{Y, attr}|}{|D_Y|}.
\end{aligned}
\end{equation}

\subsection{Bootstrap Analysis}

Experiments were conducted on MIMIC-CXR-JPG~\cite{johnson2019mimic}, a large multimodal chest X-ray dataset with images, reports, and metadata across 14 pathologies. Embeddings were extracted using BioMedCLIP~\cite{zhang2023biomedclip}.
For each setup, 100 bootstrap iterations were run by resampling training sets (1000) and test sets (300) while keeping the target class $y$, error attribute $\mathit{attr}$, error strength, and underperforming proportion fixed. To avoid inflated performance, all test sets contained 20\% underperforming samples. Each iteration trained a ResNet-18 from scratch with normalization, random flips, and ±10° rotations. Model validity required at least a 10\% accuracy gap between groups with and without the predefined attribute. We set $\gamma=10$, limited slices to five, and reported the one with the highest Precision@10~\cite{eyuboglu2022domino}. The top five distinctive tokens were recorded.

As a baseline, we used a \textit{Global TF-IDF Analysis} that treated all test samples as one group without slice discovery, to assess whether discovery yields purer error slices for auditing. Without clustering, we reported the proportion of misclassified samples containing the ground-truth error attribute as a global precision-like metric. We then compared \textit{image-only}, \textit{unimodal}, and \textit{multimodal} embeddings,where the image-only setting followed the original SDM design and served as the main reference.

The pipeline was evaluated under three common failure modes, as described in Section~\ref{failure}, following the setups below. 
For \textit{spurious correlation}, the failure model was a \textbf{pneumothorax classifier} with the error attribute \textbf{supporting devices}, trained with an induced correlation of $\rho=0.7$ between the negative class ($y=0$) and device presence ($attr=1$). 
For \textit{rare-slice undertraining}, the failure model was a \textbf{cardiomegaly classifier} with the error attribute \textbf{lateral view}, where positive samples ($y=1$) in the lateral view ($attr=0$) were underrepresented ($R(1,0)=0.02$). 
For \textit{noisy-label injection}, the failure model was a \textbf{finding classifier} with the error attribute \textbf{frontal view}, where 30\% of positive samples ($y=1$, $attr=1$) were randomly flipped to simulate label noise.

\section{Results and Discussions}

\begin{table}[h!]
\centering
\small
\begin{tabular}{l ccc}
\hline
\textbf{Embedding Type} &
\shortstack{\textbf{P@10}\\\textbf{(Corr)}} &
\shortstack{\textbf{P@10}\\\textbf{(Rare)}} &
\shortstack{\textbf{P@10}\\\textbf{(Noisy)}} \\
\hline
Image only              & 0.567 & 0.621 & 0.667 \\
Image + Text            & 0.565 & 0.696 & 0.649 \\
Image + Text + Metadata & 0.568 & 0.742 & 0.513 \\
Report Text             & 0.578 & 0.736 & 0.687 \\
Metadata only           & 0.538 & \textbf{0.909} & 0.364 \\
Report + Metadata       & 0.551 & 0.852 & \textbf{0.744} \\
Image + Metadata        & \textbf{0.638} & 0.761 & 0.687 \\
Baseline                & 0.472 & 0.301 & 0.672 \\
\hline
\end{tabular}
\caption{Mean \textit{Precision@10} across embeddings under three failure scenarios: correlation, rare slice, and noisy labels.}
\label{tab:embedding_three}
\end{table}

Table \ref{tab:embedding_three} presents the average \textit{Precision@10} scores for all three types of model failures across bootstrap iterations for each embedding type, along with the baseline. For \textit{spurious correlation}, all embedding configurations outperform the baseline (0.47), suggesting that the SDM can be effectively extended beyond image-only embeddings to other modalities.  
Among them, the best performance is achieved by the \textit{Image + Metadata} setting (0.64), which is about 15\% higher than the \textit{Image Only} setting (0.57).  
This indicates that multimodal embeddings provide complementary information that enhances error-slice discovery.  For token analysis, across all iterations, the identified tokens for each embedding setting were consistently related to supporting devices, such as ``tube'' and ``line'', showing that the token analysis effectively captures clinically relevant failure attributes.  Similar patterns were observed in the \textit{rare-slice undertraining} experiment, where all embeddings that included metadata achieved higher scores (0.74-0.91).  This result is reasonable, as the predefined error attribute (view position) is explicitly recorded in the metadata.  
In these runs, the token ``lateral'' appeared as the most distinctive word across bootstrap iterations.  

From these two experiments, we observe that modalities other than images can achieve comparable auditing performance. Since image processing is computationally expensive, the results highlight the potential of text-based modalities as efficient alternatives.

\begin{figure}[h!]
    \centering
    \includegraphics[width=0.99\linewidth]{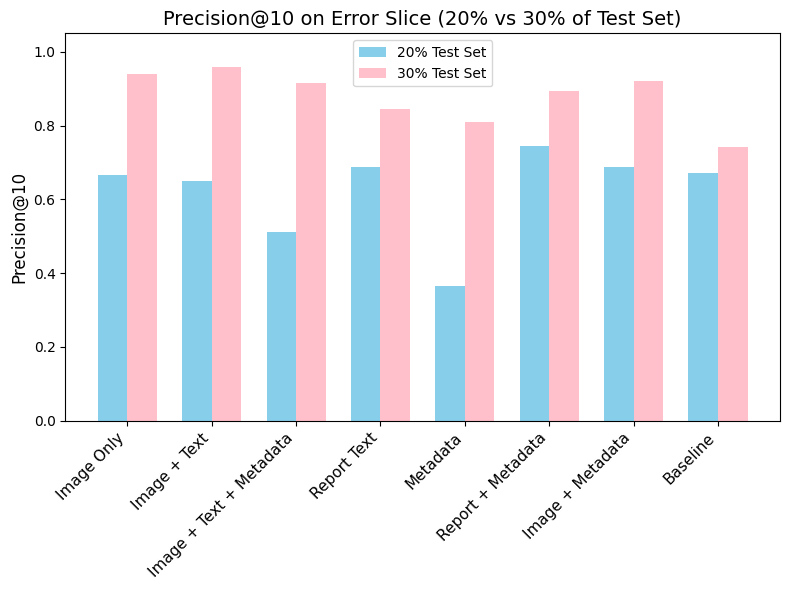}
    \caption{Average \textit{Precision@10} under noisy-label settings with 20\% and 30\% underperforming samples in the test set.}
    \label{fig:exp2_30}
\end{figure}

For \textit{noisy-label injection}, the \textit{Report + Metadata} embedding achieves the highest \textit{Precision@10} (0.744). However, overall performance in the noisy-label slice is noticeably lower than in the other two experiments, with only half of the embeddings exceeding the baseline, indicating that noisy labels pose the greatest challenge. \textit{Metadata} embeddings, which performed best in the rare-slice case, showed the worst stability here, as more than half of the bootstrap runs produced no valid slices. 
This instability results from GMM’s preference for larger clusters and Domino’s emphasis on high-error groups. Label noise increases error rates for both positive and negative samples. However, since the predefined underperforming group ($y=1$, $attr=1$) constitutes only 20\% of the test set and poorly performing negative samples dominate, the model tends to form unstable clusters around the few positive instances, resulting in incomplete auditing. As shown in Figure~\ref{fig:exp2_30}, when the proportion of underperforming samples was increased from 20\% to 30\%, all embeddings improved significantly, confirming that the pipeline remains effective but is constrained by limited test data.

For failure explanation, even in the 20\% noisy-label setting, SDMs show clear advantages over the baseline: token analysis reveals shared failure patterns across positive and negative samples. Across iterations, ``portable'' consistently emerges as a distinctive token for all embeddings, whereas the baseline yields only the generic ``normal''. Although seemingly unrelated, in practice it indicates anterior–posterior (frontal) X-rays, the true error attribute.

To further refine the pipeline, we modified the pipeline by applying Gaussian Mixture Model (GMM) clustering exclusively to misclassified samples within each class, rather than using the label-aware Domino approach on the entire test set. This design focuses clustering purely on feature representations of the samples. Preliminary experiments conducted under a noisy-label setting (one iteration) show that the \textit{Precision@5} increased by more than 100\% across 5\% underperforming groups for all embedding types except \textit{metadata-only}, suggesting a promising future direction for improving pipeline robustness.

\section{Conclusion and Future Directions}
We have presented a novel, fully-automated auditing framework for black-box medical image classifiers. To our knowledge, this is the first work leveraging the multimodal nature of medical data for model auditing. The framework effectively identifies systematic errors across diverse failure modes, showing that multimodality enhances both error detection and explanation. Future work can address the data sparsity issues observed under noisy-label settings. 
One potential improvement is to apply GMM clustering only to misclassified samples within each class, focusing the discovery purely on sample features. 
In addition, more advanced fusion strategies should be explored beyond simple concatenation to reduce potential information loss and enhance the representation quality of multimodal embeddings.

\section{Compliance with Ethical Standards}
This study used publicly available human imaging data from MIMIC-CXR-JPG \cite{johnson2019mimic}, following PhysioNet guidelines. The required ethical training was completed.

\section{Acknowledgments}
AEF acknowledges support from UKRI CDT in Artificial Intelligence for Healthcare (Grant Number: EP/S023283/1). KKB is a Founder and Director of Metalynx Ltd. trading as Aival. For the purpose of open access, the authors have applied a Creative Commons Attribution (CC BY) license to any Author Accepted Manuscript version arising.
\bibliographystyle{IEEEbib}
\bibliography{refs}

\end{document}